\documentclass[letterpaper, 10 pt, conference]{ieeeconf}  
\IEEEoverridecommandlockouts                              

\usepackage[table]{xcolor}
\usepackage{graphics} 
\usepackage{epsfig} 
\usepackage{times} 
\usepackage{amsmath} 
\usepackage{amsfonts}

\usepackage{booktabs}
\usepackage{multirow}
\usepackage{stfloats}
\usepackage{subcaption}
\usepackage{microtype}
\usepackage{float}
\usepackage{hyperref}
\usepackage{fancyhdr}
\usepackage{svg}

\title{\LARGE \bf
6-DoF Object Tracking with Event-based Optical Flow and Frames
}

\author{Zhichao Li$^{1,2,3}$, Arren Glover$^{2}$, Chiara Bartolozzi$^{2}$, and Lorenzo Natale$^{1}$
\thanks{$^{1}$Humanoid Sensing and Perception, Istituto Italiano di Tecnologia, Italy {\tt\small \{first.last\}@iit.it}}%
\thanks{$^{2}$Event-driven Perception for Robotics, Istituto Italiano di Tecnologia, Italy {\tt\small \{first.last\}@iit.it}}%
\thanks{$^{3}$University of Genoa, Genoa, Italy}
}

\begin{document}

\maketitle
\thispagestyle{empty}
\pagestyle{empty}

\begin{abstract}

Tracking the position and orientation of objects in space (i.e., in 6-DoF) in real time is a fundamental problem in robotics for environment interaction. It becomes more challenging when objects move at high-speed due to frame rate limitations in conventional cameras and motion blur. Event cameras are characterized by high temporal resolution, low latency and high dynamic range, that can potentially overcome the impacts of motion blur. Traditional RGB cameras provide rich visual information that is more suitable for the challenging task of single-shot object pose estimation. In this work, we propose using event-based optical flow combined with an RGB based global object pose estimator for 6-DoF pose tracking of objects at high-speed, exploiting the core advantages of both types of vision sensors. Specifically, we propose an event-based optical flow algorithm for object motion measurement to implement an object 6-DoF velocity tracker. By integrating the tracked object 6-DoF velocity with low frequency estimated pose from the global pose estimator, the method can track pose when objects move at high-speed. The proposed algorithm is tested and validated on both synthetic and real world data, demonstrating its effectiveness, especially in high-speed motion scenarios. 
\end{abstract}

\section{INTRODUCTION}
In computer vision and robotics, tracking object pose in 6-DoF is crucial for downstream tasks, such as object position perception, object manipulation and human-robot interaction.  6-DoF pose tracking refers to the continuous monitoring position and orientation of the object for which there is an extensive body of work and numerous of methods have been proposed, including traditional filter based methods~\cite{7487184,6696810}, deep learning strategies~\cite{9341314, wen2024foundationpose}, and hybrid solutions~\cite{9568706, 10342300, 9363455}. However, most of these methods are only suitable for low-speed object motion, partly due to camera limitations at high-speed~\cite{9568706, 10342300}. 6-DoF object pose tracking for high-speed motion object is still an open problem. 
Frame-based optical flow information has been used to enhance object pose tracking for high-speed motion scenarios~\cite{9568706}. However, frame-based optical flow is also not reliable when an object moves at high-speed due to the latency introduced by the frame rate and to the motion blur originating from the process of acquisition of complete frames. 

Event cameras are promising for solving problems in fast motion scenarios due to their unique characteristics, including low latency, high dynamic range and asynchronous event triggering~\cite{4444573, gallego2020event}. The temporal integration of light can be much faster than traditional sensors, thereby avoiding motion blur.
The advantages of event cameras make them a promising vision sensor for tackling problems related to high-speed motion during object pose tracking. 


Despite event cameras having many advantages, algorithms utilizing their output must be customized to effectively extract the information and maintain high-frequency. For example, event cameras respond to brightness change, i.e. a measure of the derivative of the pixel intensity that traditional sensors measure. Event cameras are therefore highly suitable together with algorithms that directly measure derivative-based signals, e.g. optical flow. Robust state estimation has instead been achieved using hybrid systems that fuse velocity estimation from event cameras and absolute object pose estimation using traditional vision sensors, e.g.~\cite{10342300}. It was shown that as little as 1-DoF measurement model was sufficient to perform camera tracking due to the precise event timing~\cite{8794255}. However, the same model adopted to object tracking had difficulty tracking along all axes simultaneously~\cite{10342300} due to the reduced amount of information and observability problems when shifting from full-scene motion, to small amounts of motion occurring locally in the sensor plane, and the self-occlusion that occurs observing rotating objects. The method could successfully measure speed along an axis but could not correct velocity direction accurately due to the low dimensional observation.
\begin{figure}[t]
    \centering
    \includegraphics[width=1.0\linewidth]{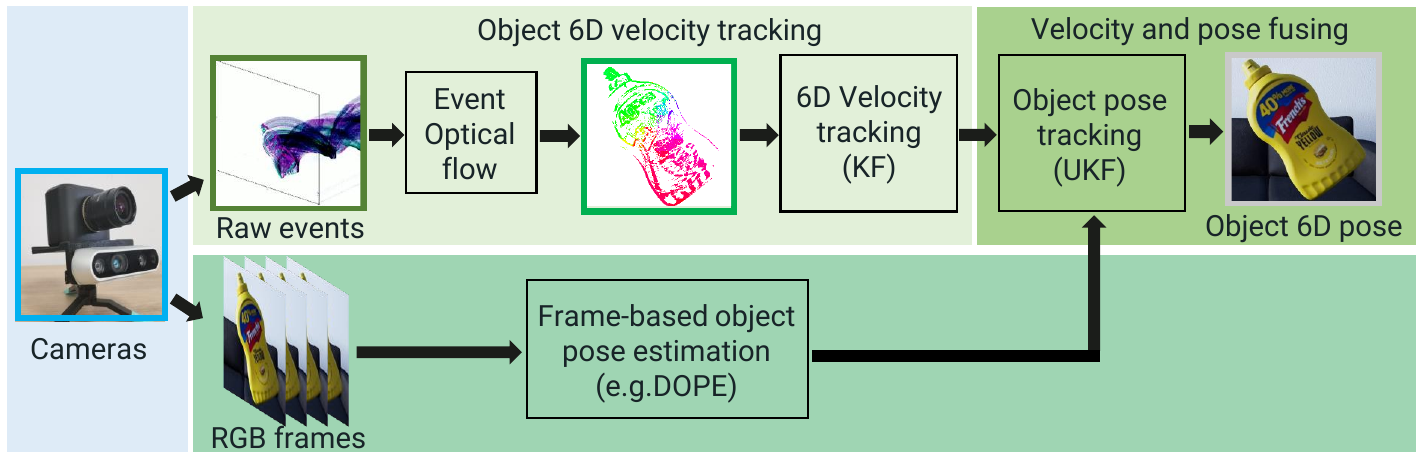}
    \caption{Pipeline of the proposed object 6-DoF pose tracker: A custom setup is made with an event camera and a RealSense. An object 6-DoF velocity tracker is implemented with a Kalman Filter (KF) taking event-based optical flow as input. RealSense provides frames for a global object pose estimator. The tracked object 6-DoF velocity and the estimated object 6-DoF pose are fused by an Unscented Kalman Filter (UKF) for object 6-DoF pose tracking.}    
    \label{pipeline}
\end{figure}

In this paper, we explore solutions to improve object pose tracking with a hybrid event camera and frame-based camera setup that exploits the advantages of both sensing techniques\,\rule[0.5ex]{0.7em}{0.55pt}\,event cameras have the capability to encode high-speed motion information while frame cameras provide rich information that is suitable for object pose estimation.

\begin{figure}[t]
    \centering
    \includegraphics[width=0.95
    \linewidth]{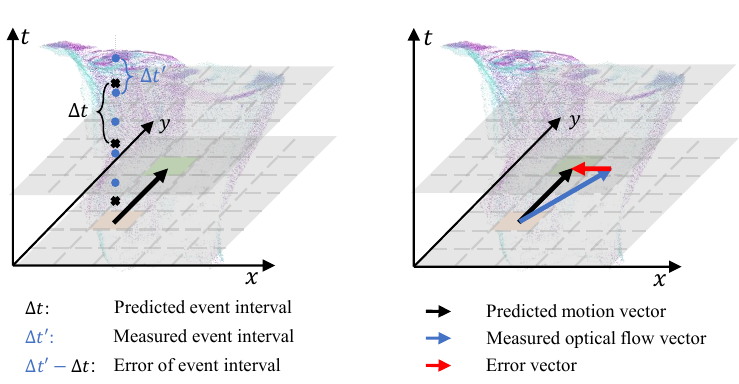}

    \caption{Difference between 1-DoF and 2-DoF error measurements. Left: The 1-DoF error is computed based on the temporal difference ($\Delta{t}$, $\Delta{t'}$) between events in the same pixel. Right: The 2-DoF error is derived from optical flow and predicted pixel motion, capturing the displacement in image space. The 2-DoF error provides richer information compared to the 1-DoF error, as it accounts for changes in both spatial direction and magnitude, rather than solely temporal variations.}    
    \label{1D2D}
\end{figure}

Specifically, a Kalman filter is adopted for object velocity tracking that uses a 2-DoF observation model based on optical flow. The tracked velocity is fused together with a frame-based object pose estimator (DOPE~\cite{tremblay2018corl:dope}) using an Unscented Kalman filter. We demonstrate that the 2-DoF observation model outperforms the 1-DoF generative event model~\cite{10342300} for the task of object tracking (Fig.~\ref{1D2D}), which also removes the dependency of the RGB camera in the object velocity pipeline. The optical flow algorithm introduced is a lightweight but effective algorithm that can produce a flow vector for each event and is scalable to real-time operation. To improve the accuracy of event optical flow, an aperture robustness flow constraint is integrated following~\cite{9268109} and a novel weighting strategy is utilized~\cite{jaegle2016fast}. In the experiments, DOPE output is set to 5 FPS in order to simulate using low power/small GPUs like the ones designed for embedded systems.
The proposed 6-DoF object tracking algorithm achieves comparable results on slow moving objects and consistently higher performance when tracking high-speed objects. 

\section{RELATED WORK}



Vision-based object pose estimation can be separated into two main groups. Classical modeling solutions typically include iterative Bayesian filters, such as the Kalman filter or the particle filter~\cite{7487184,6696810}. The advantage of this type of method is that algorithms are lightweight thus have a low computational burden. However, it is not always simple to accurately model all dynamics and external disturbances that are needed for accurate results. 
Instead data-driven approaches, e.g. Deep Neural Networks (DNN), learn non-linear solutions from data and can therefore account for difficult to model phenomena~\cite{9341314, wen2024foundationpose}. These approaches are dependent on good training data, and are heavier as the exact number of layers and connections required to represent the underlying functions are over-represented in the network architecture. 

Both classical and data-driven object tracking methods that use frame-based cameras as input drop in accuracy with high-speed motion due to image blur resulting from camera exposure time. When severe image blur exists, the extracted motion and pose information is no longer accurate, which affects algorithm performance and even causes failures. In addition, high-speed motions result in large jumps between frames in which object pose estimation is unknown. Training for these situations goes to some lengths to improve performance but doesn't address the root cause of poor input data because of sensor unsuitability.

Optical flow is basic vision information that encodes pixel motion caused by object and scene motions relative to the camera and can be integrated in object tracking algorithms. In ROFT~\cite{9568706}, a frame-based optical flow-aided strategy was utilized and combined with Kalman filtering to enhance a state-of-the-art DNN object pose estimator DOPE~\cite{tremblay2018corl:dope}, ultimately, improving pose tracking performance for high-speed motions.
Classic optical flow algorithms include the Lucas-Kanade method~\cite{lucas1981iterative} and the Horn-Schunck optical flow algorithm~\cite{horn1981determining}. Subsequently, DNN based optical flow algorithms were introduced, one of representative work is RAFT~\cite{teed2020raft}. 

Although optical flow can estimate high-speed motion, limitations are imposed by the sensor technology, such as motion blur. Event cameras have a natural advantage in encoding optical flow information due to their unique characteristics, of which some improve robustness to motion blur~\cite{gallego2020event}. They have potential in many applications, such as autonomous driving~\cite{Gehrig21ral} and object tracking~\cite{10342300, 10611511}. Event-based optical flow algorithms can also be divided into two types: decoding optical flow by analyzing spatio-temporal information among events, such as plane fitting~\cite{benosman2012asynchronous}, surface matching~\cite{nagata2021optical} and triplet matching~\cite{shiba2022fast}. Another type is to learn event feature for optical flow estimation relying on convolution or spiking neural networks~\cite{zhu2018ev, paredes2019unsupervised}. Optical flow from events has been adopted in object pose tracking~\cite{liu2024optical} and camera pose estimation~\cite{ren2024motion}. 


\section{METHODS}\label{sec:method}

The proposed method uses RGB frames $\{I_i\}$, synchronized depth images $\{D_i\}$, and a high temporal precision event stream~$\{E_q\}$. Our target is to track object 6-DoF pose even during high-speed motions. An event-by-event optical flow algorithm (i.e. at the same temporal precision as the raw event-stream) is utilized for object motion measurement on the visual plane. Event optical flow is accumulated spatially in~$\{F_i\}$. An object pose estimator provides estimated object pose $\{T_m\}$, at a low rate due to computational limitations. Our pipeline, shown in Fig.~\ref{pipeline}, consists of three parts: 1) event-based optical flow estimation, 2) object 6-DoF velocity tracking based on the event-based optical flow using a Kalman Filter, and 3) object 6-DoF pose estimation by fusing velocity and low frequency pose observations using an Unscented Kalman filter.

\subsection{Event-based optical flow}
Optical flow is calculated from the event stream~$\{E_q\}$, formed by brightness changes triggering pixels, caused by the motion of 3D points of an object moving in the scene. 
By decoding event optical flow information, combined with interaction matrix~(equation.~(\ref{eq:ekf_interaction_model}), see Sec.\ref{Sec:implicit_measurement_model}), the linearized relationship between the 3D scene motion and 2D optical flow can be built. Therefore, 3D object motion information can be estimated by analyzing object optical flow information. 

In this work, event optical flow is calculated by analysing the spatio-temporal relationship between individual events, particularly as an object moves in space (3D), the trajectory of points on the object sweeps across the 2D image plane. By considering a constant velocity over small regions of the image plane, then three neighbouring events $\boldsymbol{e}_i$, $\boldsymbol{e}_j$ and $\boldsymbol{e}_k$ on the 2D pixel plane satisfy the following spatio-temporal constraint~\cite{shiba2022fast}:
\begin{equation}
    \begin{split}
        \frac{\boldsymbol{x}_j - \boldsymbol{x}_i}{t_j-t_i}  = \frac{\boldsymbol{x}_k - \boldsymbol{x}_j}{t_k-t_j + \xi}\\
    \end{split}
\end{equation}%
where $\boldsymbol{x}_i$, $\boldsymbol{x}_j$ and $\boldsymbol{x}_k$ are 2D pixel coordinates, $(u, v)$, of the three events, $t_i$, $t_j$ and $t_k$ are their respective timestamps, and $\xi$ is a tolerance parameter. If $\boldsymbol{e}_i$, $\boldsymbol{e}_j$ and $\boldsymbol{e}_k$ can  satisfy the condition, then optical flow of pixel $\boldsymbol{x}_i$ at timestamp $t_i$ can be calculated as:
\begin{equation}\label{eq:flow_calculation}
    \begin{split}
        \mathcal{F}(\boldsymbol{x}_i) = (\boldsymbol{x}_k – \boldsymbol{x}_i)/(t_k-t_i)\\
    \end{split}
\end{equation}%

However, the above spatio-temporal constraint alone does not guarantee that the triplet events belong to the same motion trajectory. Incorrect matches can still be found, and eventually introduce errors into the 6-DoF object velocity estimation.

To mitigate the impact of erroneous flow vectors, we use a spatio-temporal registration strategy. Specifically, the image plane is split into small grids, each covering a region of interest~(\textit{RoI}). Given the assumption of optical flow consistency in small \textit{RoIs}~\cite{lucas1981iterative}, all searched event triplets within each \textit{RoI} are used to estimate a single flow vector using the following cost function.
\begin{equation}\label{eq:least-square}
        \mathcal{F} = \arg\min_{\mathcal{F}' \in \mathcal{M}}\bigg(\sum\limits_{i=1}^{n}\min_{\mathcal{F}_{ic}\in M_i}\mid \mathcal{F}'-\mathcal{F}_{ic}\mid \bigg)
\end{equation}%
where $\mathcal{F}$ is the target optical flow, which minimizes the cost function, $\mathcal{M}$ represents candidate flow vectors calculated by the found triplets following Equation~(\ref{eq:flow_calculation}) within $n$ events in the $\textit{RoI}$, and $M_i \subset \mathcal{M}$ covers all candidate flows searched for the single event $\boldsymbol{e}_i$.

\subsection{Object velocity estimation}
\label{Sec:object_velocity_estimation}
A Kalman filter is used to combine event-based optical flow, $\{{F}_i\}$, accumulated spatially, and the observed depth, $\{D_i\}$, to calculate the 6-DoF object velocity.

\subsubsection{State definition}
The object 6-DoF velocity includes linear and rotation velocity along each axis, and is defined as:
\begin{equation}\label{eq:ekf_state}
    \begin{split}
        \mathcal{V}_i=[v_{oi}, \omega_i]
    \end{split}
\end{equation}
where, $v_{oi} \in \mathbb{R}^3$ represents the velocity of a point that instantaneously coincides with the origin of the camera and moves as if it was rigidly attached to the object. $\omega_i \in \mathbb{R}^3$ is the angular velocity of the object with respect to the same point.

\subsubsection{Motion model}
To maintain applicability to a wide range of tracking tasks we do not impose a particular motion model on the object, and instead assume a decaying velocity model, with adjustment parameter $\alpha$. 
\begin{equation}\label{eq:ekf_motion_model}
	\begin{split}
		\mathcal{V}_{i+1} &= \alpha \mathcal{V}_{i} + w,\\
		w &\sim \mathcal{N} \left(0, \mathrm{diag}(Q_{v}, Q_{\omega}) \right)
	\end{split}
\end{equation}
where the propagation noise, $\omega$, follows a zero-mean Gaussian distribution with covariance $Q_{v} \in \mathbb{R}^{3\times 3}$ and $Q_{\omega} \in \mathbb{R}^{3\times 3}$ associated to the linear and angular velocity, respectively. When $\alpha = 1$, a constant velocity model is realized, instead setting $\alpha \in (0, 1)$ achieves a counteraction to the phenomenon that events only occur during motion. In this work, $\alpha$ is set to 0.5. The scenario in which no events are produced indicates the object is stationary, but as no data is produced, no correction step can be performed. Instead, the Kalman prediction step must be used to achieve a zero velocity.

\subsubsection{Event optical flow based measurement model}
\label{Sec:implicit_measurement_model}
The measurement model uses the optical flow estimation, and to ensure observability, it is essential to collect a sufficient number of event optical flow measurements. We define the collected event optical flow measurements as a vector:
\begin{equation}\label{eq:ekf_measurement}
	\begin{split}
		F(u, v)_i = 
        \begin{bmatrix} 
        ... \\ 
        \mathcal{F}(u,v) \\ 
        ... 
        \end{bmatrix}
	\end{split}
\end{equation}
each $(u,v)$ pixel location for which there occurred an event and event optical flow is measured since the last Kalman correction. 
The proposed method relies on the pinhole camera model and the time derivative form. Therefore the estimated object 6-DoF velocity $\mathcal{V}_i$ can be mapped to 2D image plane of event optical flow.    \begin{equation}\label{eq:ekf_interaction_model}
	\begin{split}
    \hat{F}(u, v)_i = 
        \begin{bmatrix} 
        ... \\ 
        \mathcal{\hat{F}}(u,v) \\ 
        ... 
        \end{bmatrix},\quad  \mathcal{\hat{F}}(u,v) = \mathcal{J}(u, v)\mathcal{V}_i\Delta_{T}
	\end{split}
\end{equation}

\begin{equation}\label{eq:ekf_interaction_model_jacobi}
	\begin{split}
    \mathcal{J}(u, v)= [\mathcal{J}_{v_o} \quad \mathcal{J}_{\omega}],  \: \mathcal{J}_{v_o} = \begin{bmatrix} 
        \frac{f_x}{d} \qquad 0 \qquad \frac{-(u-c_x)}{d}  \\ 
        0 \qquad \frac{f_y}{d} \qquad\frac{-(v-c_y)}{d}\\ 
        \end{bmatrix}\\
        \mathcal{J}_{\omega} = \begin{bmatrix} 
        \frac{-(u-c_x)(v-c_y)}{f_y} \qquad \frac{f_x^2+(u-c_x)^2}{f_x} \qquad\frac{-(v-c_y)f_x}{f_y} \\ 
        \frac{f_y^2+(v-c_y)^2}{f_y} \qquad \frac{(u-c_x)(v-c_y)}{f_x} \qquad \frac{(u-c_x)f_y}{f_x}
        \end{bmatrix}
	\end{split}
\end{equation}
where $d\in\mathbb{R}$ is the depth associated to the pixel $(u, v)$, and $c_x$, $c_y$, $f_x$ and $f_y$ are camera intrinsic parameters. For the derivation of $\mathcal{J}_{\omega}$ refer to~\cite{piga2022hybrid}.

\subsubsection{Normal flow constraint}
\label{sec:normal_flow_constraint}
 During optical flow estimation, the aperture problem results in the measured flow calculated as normal to the moving edge if not enough unique features are present in the \textit{RoI}. Missing the unobservable flow component parallel to the edge direction leads to imprecise 6-DoF object velocity estimation. Since, the observed flow $\mathcal{F}(u,v)$ in the Equation~(\ref{eq:ekf_measurement}) captures only a component of the full motion of the estimated object movement, a geometric constraint can be introduced to relate it to the expected flow $\hat{\mathcal{F}}(u,v)$, in the Equation~(\ref{eq:ekf_interaction_model}).
 Following~\cite{9268109}, the measurement model is therefore constructed as:
\begin{equation}\label{eq:ekf_measurement_model}
	\begin{split}
		Z_i = 
        \begin{bmatrix} 
        ... \\ 
        r_k + \mathcal{\nu} \\ 
        ... 
        \end{bmatrix}
	\end{split}
\end{equation}
$r_k = \frac{\mathcal{F}_k \cdot \mathcal{\hat{F}}_k}{||\mathcal{F}_k||} \cdot \frac{\mathcal{F}_k}{||\mathcal{F}_k||} - \mathcal{F}_k$ is the residual, and $\nu\sim\mathcal{N} \left(0, R_{F} \right)$ is the noise.


\subsubsection{Measurement weighting}
In order to further improve the measurement accuracy for object tracking optical flow observations with large errors are considered outliers and have less contribution to the 6-DoF velocity estimation. Valid measurements are assigned higher weights, while potential lower-quality measurements are given lower weights. 

Given a set of measurement residual $\{r_k\}$ from Equation~(\ref{eq:ekf_measurement_model}). Weights are then derived as:
\begin{equation}
    \begin{split}
        L_k = max(\frac{1}{2b}e^{-\frac{|{r}_k-M|}{b}}, 1e^{-6})
    \end{split}
\end{equation}
where $b$ is Laplacian coefficient, and $M$ is the median of measurement residual $\{{r}_k\}$~\cite{jaegle2016fast}. 

\subsubsection{Kalman Filter}
By combining the propagation model Equation~(\ref{eq:ekf_motion_model}) and the proposed measurement model Equation~(\ref{eq:ekf_measurement_model}) a standard Kalman filter is implemented to track the object velocity. The state $\mathcal{V}_{k}$ is predicted through the propagation model, and during the correction step, the measurements are incorporated to update the state.

\subsection{Object pose tracking}
\label{Sec:object_pose_tracking}
In this section, we detail the filtering strategy of combining low frequency pose measurements, $\{T_m\}$, for which many off-the-shelf solution exist using traditional computer vision~\cite{tremblay2018corl:dope}, and high-frequency 6-DoF velocity estimates, $\{\mathcal{V}_i\}$, to realize robust 6-DoF object pose tracking. 

\subsubsection{State definition}
We define the object state $x_i$ that needs to be tracked over time as
\begin{equation}
	x_i = [t_i, q_i]^T
\end{equation}
where $t_i \in \mathbb{R}^{3}$ is the Cartesian position and $q_i \in \mathbb{H}$ is a unitary quaternion representing the 3D orientation of the object. An alternative approach is to incorporate the object 6-DoF velocity $\mathcal{V}_i$ as part of the state, as in~\cite{9568706}.
\subsubsection{Motion model}
The object pose propagates according to the estimated velocity, $\mathcal{V}_i$: 
\begin{equation}\label{eq:ukf_motion_model}
\begin{split}
    x_{i+1} =
    \begin{bmatrix}
        t_{i+1}\\
        q_{i+1}
    \end{bmatrix}
    &= 
    \begin{bmatrix}
        \left( I_{3} + \hat{\omega}_{i} \Delta_{T} \right) t_{i} + v_{oi} \Delta_{T}\\
        F(\omega_{i}) q_{i}\\
    \end{bmatrix} \oplus w,\\
    w &\sim \mathcal{N}(0, \mathrm{diag}(Q_{t},
    Q_{q})).
    \end{split}
\end{equation}
where $v_{oi}$ and $\omega_i$ are measured linear and angular velocity in Equation~(\ref{eq:ekf_state}). $\hat{\omega}_{i}$ represents the skew-symmetric matrix, and $\Delta_{T}$ is the velocity measurement interval time. The term $\hat{\omega}_{i} \Delta_{T} t_{i}$ represents the reflected linear motion from rotation at the origin of the camera. It combines the linear motion with translation velocity of time to obtain the position of the next timestamp.
$F(.)$ is the standard quaternion kinematics transition matrix \cite{chiella2019quaternion}, and $\oplus$ represents the operation of adding noise $w$ taking into account the quaternion arithmetic \cite{chiella2019quaternion}. $Q_{t} \in \mathbb{R}^{3 \times 3}$, $Q_{q} \in \mathbb{R}^{3 \times 3}$ are the noise covariance matrix for the translational and rotational components of the state, respectively.

\subsubsection{Measurement model}
The estimated 6-DoF object pose is utilized for state correction:
\begin{equation}
    y_{i}(T_{i}) = 
    \begin{bmatrix}
    p(T_{i}) \\
    q(T_{i}) \\
    \end{bmatrix},
\end{equation}
where $p(T_{i}) \in \mathbb{R}^{3}$, $q(T_{i}) \in \mathbb{H}$ are the Cartesian position and the quaternion components of the measured pose $T_{i}$.  The measurement model relating $y_{i}$ with the state vector $x_{i}$ is expressed as
\begin{equation}\label{eq:ukf_measurment_model}
\begin{split}
    y_{i}(x_{i}) &= 
    \begin{bmatrix}
    t_{i}\\
    q_{i}
    \end{bmatrix}
    \oplus \nu,\\
    \nu &\sim \mathcal{N}(0, \mathrm{diag}(R_{t}, R_{q})),
    \end{split}
\end{equation}
with $R_{t} \in \mathbb{R}^{3 \times 3}$ and $R_{q}$ the noise covariance matrices associated to the translational and rotational components of the state. If pose measurements are not available at time $i$, the state correction step of the UKF is skipped.

\subsubsection{Unscented Kalman filtering}
Since the state is represented in quaternion form and the motion model is non-linear, the state $x_{i}$ is tracked using a quaternion-based Unscented Kalman Filter (UKF)~\cite{chiella2019quaternion}.

\section{EXPERIMENTS}
We conduct experiments to validate the effectiveness of our method through qualitative and quantitative analyses using synthetic and real-world datasets. Comparisons are performed against SOTA RGB-D trackers ROFT~\cite{9568706} and se(3)-TrackNet~\cite{9341314}, particularly in high-speed motion scenarios.

\begin{figure}[t]
    \centering
    \includegraphics[width=0.99
    \linewidth]{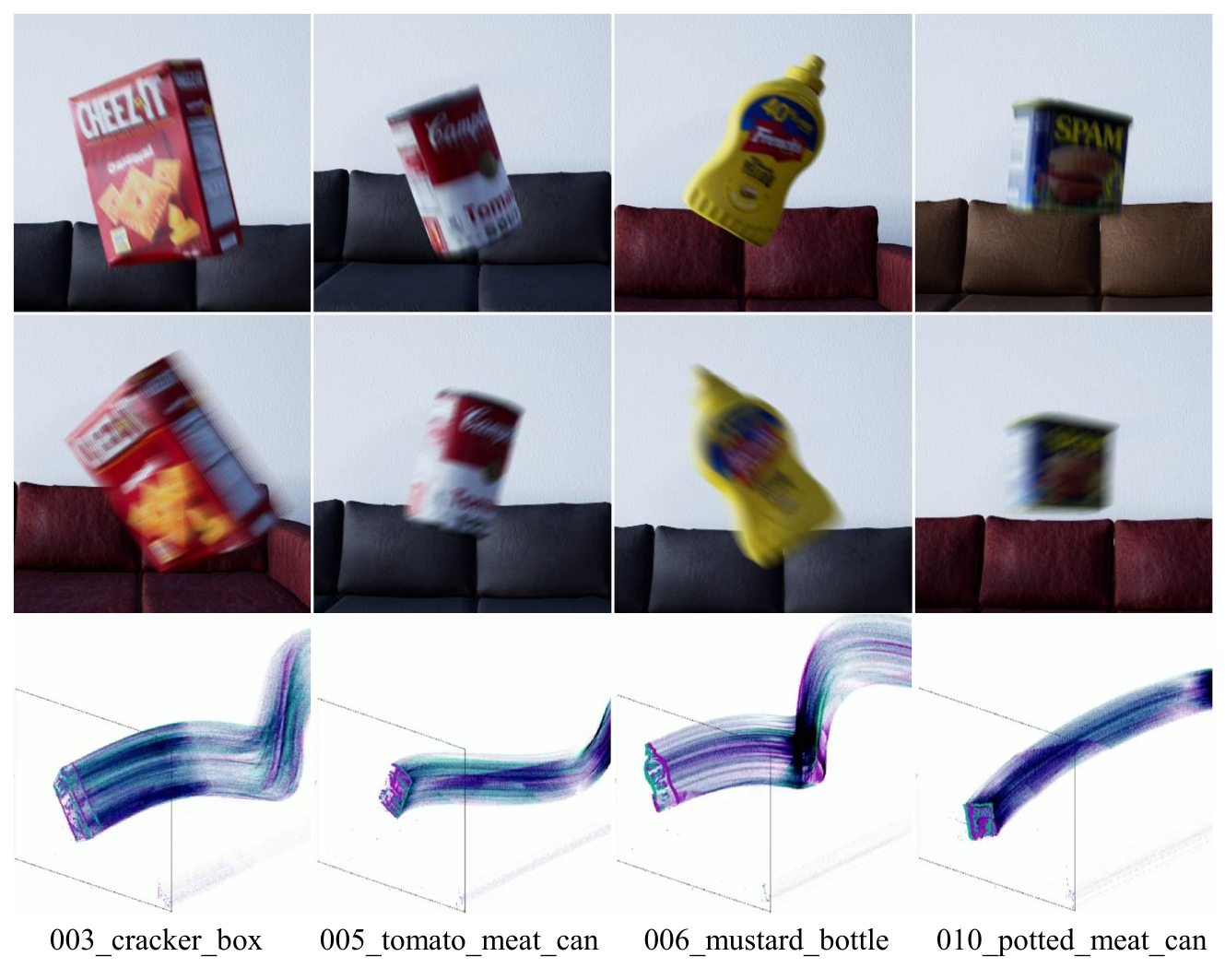}

    \caption{{The objects used in the synthetic datasets under (top row) regular motion and (middle row) fast motion. Motion blur is observable in the fast-motion snapshots. Events streams are visualized in the last row. Due to the large temporal resolution, there is almost no image blur in the event representation.}}    
    \label{motion_blur_snapshot}
\end{figure}

\subsection{Testing data sequences} \label{Testing data}
\begin{figure}[t]
    \centering
    \includegraphics[width=0.9
    \linewidth]{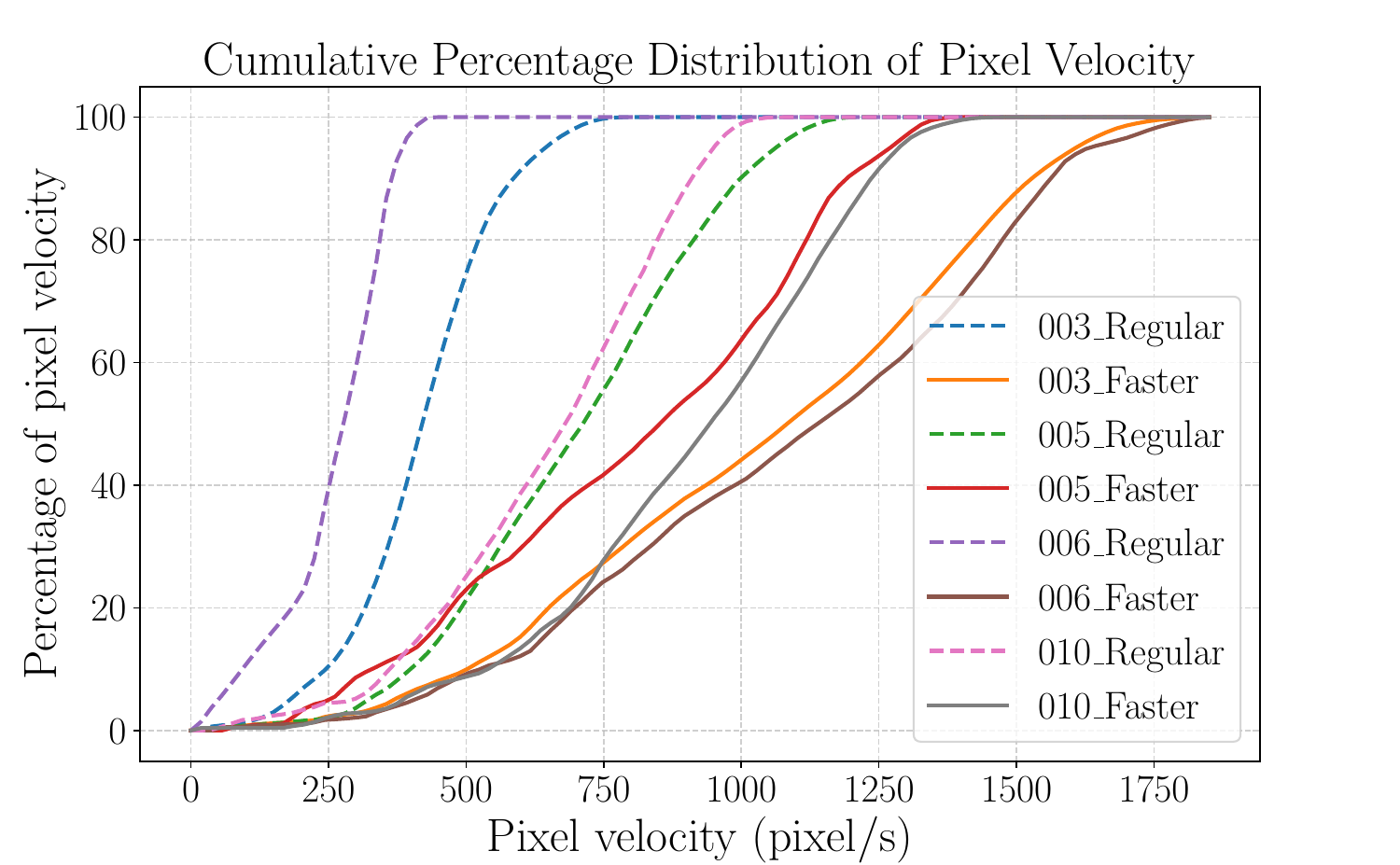}

    \caption{{Cumulative percentage distribution of pixel velocity for testing sequences.}}    
    \label{cumulative_percentage_distribution}
\end{figure}

A synthetic dataset is generated using Unreal Engine with YCB object set~\cite{7251504}. Four different objects with different sizes and shapes are chosen including 003\_cracker\_box, 005\_tomato\_soup\_can, 006\_mustard\_bottle and 010\_potted\_meat\_can. Each object randomly follows 6-DoF motion trajectories with both regular and fast velocities with velocity distributions as shown in Fig.~\ref{cumulative_percentage_distribution}. RGB frames are initially generated at 500~FPS and used as input to an event-camera model for events generation. These frames are then averaged within fixed time periods to achieve 60~FPS. The averaging of multiple frames over time simulated motion blur as shown in Fig.~\ref{motion_blur_snapshot}. Ground truth object pose and object 6-DoF velocity are extracted directly from the simulator. A real-world dataset was collected using a calibrated pair of RealSense D415 and an event camera. The RealSense achieves 60~FPS with 640$\times$480 resolution, and the event camera is asynchronous with a 640$\times$480 resolution. Two YCB objects, 003\_cracker\_box and 006\_mustard\_bottle, are moved as fast as possible in the field of view of the cameras by a human operator. The real datasets do not have ground truth poses and are used to achieve qualitative results, but with real sensors.

\subsection{Evaluation metric} \label{evaluation metric}

Tracking root mean square error~(RMSE) is calculated with respect to ground truth pose, position errors are calculated in Euclidean distance, and rotation errors are presented as rotation degree vector in $\mathfrak{se}(3)$ space. In the qualitative analysis, tracking curves of different algorithms are visualized for comparison on synthetic data sequences, while for real world data sequences, the rendered object using tracked pose is visualized on top of the image.   

\begin{table*}[pt!]
    \vspace{0.7em}
    \renewcommand\tabcolsep{2.3pt}
	\centering
    \scriptsize
    \scalebox{1.1}{
    \begin{tabular}{c| c |c c| c c| c| c c| c c}
		\cline{1-9} 
        \toprule[1.1pt]
		\multicolumn{2}{c}{\textbf{Sequence}}                                        
		&\multicolumn{2}{c}{\textbf{Regular}}  & \multicolumn{2}{c}{\textbf{Faster}}  &\multicolumn{1}{c}{\textbf{}}  &\multicolumn{2}{c}{\textbf{Regular}}  & \multicolumn{2}{c}{\textbf{Faster}}  \\
		\hline
        Obj.  & Methods    	& $e_{p}(\sigma)(cm)$     & $e_{r}(\sigma)(deg)$   & $e_{p}(\sigma)(cm)$     & $e_{r}(\sigma)(deg)$ &Obj. & $e_{p}(\sigma)(cm)$     & $e_{r}(\sigma)(deg)$   & $e_{p}(\sigma)(cm)$     & $e_{r}(\sigma)(deg)$\\
		\hline
		\multirow{3}{*}{003} 
        &\multicolumn{1}{c|}{Ours\,(Velocity)}       
		&\cellcolor[gray]{0.95}7.20(3.32)  &15.36(4.77) &\cellcolor[gray]{0.95}6.49(2.50)  &26.81(11.22) &\multirow{3}{*}{005} &\cellcolor[gray]{0.95}12.46(6.68)   &20.37(7.12)  &\cellcolor[gray]{0.95}14.46(7.14)  &22.08(11.37) \\

        &\multicolumn{1}{c|}{Ours\,(Pose)}       
		&\cellcolor[gray]{0.95}4.80(2.63)  &7.98(3.87) &\cellcolor[gray]{0.95}14.17(8.48) &\textbf{7.68}(4.34)  & &\cellcolor[gray]{0.95}4.18(2.10)  &9.39(3.87)     &\cellcolor[gray]{0.95}8.07(4.50)  &10.85(4.97)\\ 

        &\multicolumn{1}{c|}{Ours\,(GE)\,\cite{10342300}}        
		&\cellcolor[gray]{0.95} 4.03(2.39)  &8.18(3.98)  &\cellcolor[gray]{0.95} 14.17(8.68) &8.74(5.78) & &\cellcolor[gray]{0.95}4.06(2.08)  &9.58(4.19)   &\cellcolor[gray]{0.95}7.70(4.37) &11.02(5.03) \\

		&\multicolumn{1}{c|}{\textbf{Ours\,(OF)}}        
		&\cellcolor[gray]{0.95}\underline{{1.95}(1.09)}  &\underline{7.30(3.66)}
        &\cellcolor[gray]{0.95}\underline{\textbf{3.30}(1.93)}  &\underline{9.13(4.96)} 
        & &\cellcolor[gray]{0.95}\underline{2.13(0.84)}  &\underline{\textbf{9.25}(4.08)}    &\cellcolor[gray]{0.95} \underline{\textbf{2.89}(1.27)} &\underline{\textbf{10.38}(4.35)} \\

        &\multicolumn{1}{c|}{ROFT\,\cite{9568706}}
		&\cellcolor[gray]{0.95} 1.54(0.40) &\textbf{2.61}(1.04) &\cellcolor[gray]{0.95} 6.76(3.21) &27.73(17.51) & &\cellcolor[gray]{0.95}\textbf{1.77}(0.63)  &9.29(3.87)   &\cellcolor[gray]{0.95}3.25(1.41) &17.40(7.58)\\ 
        
        &\multicolumn{1}{c|}{se(3)-TrackNet\,\cite{9341314}}
		&\cellcolor[gray]{0.95} \textbf{0.86}(0.52) &5.20(3.32)  &\cellcolor[gray]{0.95} 24.13(9.62)  &29.90(11.86) & &\cellcolor[gray]{0.95} 28.27(8.81)  &34.90(9.41)  &\cellcolor[gray]{0.95} 26.04(9.67) &22.48(8.40)\\
		\hline
        \\
        
		\hline
		\multirow{3}{*}{006} 
        &\multicolumn{1}{c|}{Ours\,(Velocity)}       
        &\cellcolor[gray]{0.95}31.13(15.56)   &17.29(7.19)  &\cellcolor[gray]{0.95}6.64(3.24)   &20.25(9.22) 
        &\multirow{3}{*}{010} 
        &\cellcolor[gray]{0.95}11.52(5.66)   &15.09(5.70)  &\cellcolor[gray]{0.95}6.15(2.98)   &16.04(6.18) \\

        &\multicolumn{1}{c|}{Ours\,(Pose)}       
		&\cellcolor[gray]{0.95}3.29(1.50) &5.89(1.73) &\cellcolor[gray]{0.95}13.84(7.64)   &11.98(5.11) &
        &\cellcolor[gray]{0.95}13.69(9.93)   &21.89(16.38)   &\cellcolor[gray]{0.95}13.58(6.71)  &16.59(7.92)\\

        &\multicolumn{1}{c|}{Ours\,(GE)\,\cite{10342300}}        
		&\cellcolor[gray]{0.95}2.19(1.03)  &5.89(1.74)   &\cellcolor[gray]{0.95}13.56(7.46)  &11.91(5.20) 
        &
        &\cellcolor[gray]{0.95}13.16(8.68)  &21.29(15.83)    
        &\cellcolor[gray]{0.95}13.74(6.63)  &15.16(6.39) \\
        
		&\multicolumn{1}{c|}{\textbf{Ours\,(OF)}}        
		&\cellcolor[gray]{0.95}\underline{1.61(0.68)}  &\underline{5.78(1.75)}    &\cellcolor[gray]{0.95}\underline{\textbf{4.66}(2.29)}  &\underline{\textbf{11.90}(4.63)} 
        &
        &\cellcolor[gray]{0.95}\underline{4.42(3.31)}  &\underline{12.96(7.99)}     &\cellcolor[gray]{0.95}\underline{\textbf{4.90}(2.13)}  &\underline{\textbf{14.82}(6.09)}\\ 
        
        &\multicolumn{1}{c|}{ROFT\,\cite{9568706}}
		&\cellcolor[gray]{0.95}1.16(0.59)  &4.88(1.62)   &\cellcolor[gray]{0.95}4.95(2.64)  &21.49(10.95) 
        &
        &\cellcolor[gray]{0.95}3.41(1.44)  &19.81(8.78)   
        &\cellcolor[gray]{0.95}74.12(35.59) &80.28(38.73) \\ 
        
        &\multicolumn{1}{c|}{se(3)-TrackNet\,\cite{9341314}}
		& \cellcolor[gray]{0.95} \textbf{0.77}(0.09)  &\textbf{2.78}(1.11)  &\cellcolor[gray]{0.95}89.82(47.29)  &87.64(49.20)
        &
        &\cellcolor[gray]{0.95}\textbf{0.84}(0.34)  &\textbf{11.37}(5.81)  
        &\cellcolor[gray]{0.95} 15.54(6.37)  &16.24(6.35) \\
        \hline

        \bottomrule[1.1pt]

	\end{tabular}}
        \caption{Tracking results of variety methods on synthetic object motion sequences. Ours\,(Velocity): Tracking only by integrating estimated object velocity~\ref{Sec:object_velocity_estimation}; Ours\,(pose): Using only low frequency\,(at 5Hz, the same input frequency used for all Ours methods) pose from DOPE with the filter~\ref{Sec:object_pose_tracking}; Ours\,(GE)~\cite{10342300} uses a generative event model on all axes simultaneously; Ours\,(OF) is our proposed method using optical flow, it achieves better performance for faster motion sequences.} \label{Main_table}

\end{table*}

\subsection{Comparison with RGB-D object pose trackers on synthetic data}\label{comparison with RGB-D object pose trackers}
In this experiment, we compare with RGB-D object pose tracker se(3)-TrackNet~\cite{9341314}. se(3)-TrackNet is a representative CNN-based object 6-DoF pose tracker,  it estimates the relative pose between the current pose, as represented by the collected RGB-D image, and the previously tracked pose.

In general, Ours\,(OF) utilizing event-based optical flow, achieves stable performance across the object speeds, compared to se(3)-TrackNet. The tracking performance of se(3)-TrackNet degrades as the motion velocity increases, as shown in Table~\ref{Main_table} and Fig.~\ref{cumulative_percentage_distribution}. 
For regular velocity motions, se(3)-TrackNet achieves better position tracking results on 003\_cracker\_box, and the smallest position and rotation errors on 006\_mustard\_bottle and 010\_potted\_meat\_can. However, it fails on the 005\_tomato\_meat\_can, which has relative higher pixel velocity. While our method obtains comparable results, and  shows better tracking performance in terms of position and rotation error on faster velocity motions, se(3)-TrackNet fails for all faster samples.

Faster motion typically results in higher pixel velocity and more severe motion blur. It requires a larger training set to cover high-speed motion scenarios, and more robust feature extractor for se(3)-TrackNet to deal with motion blur. On the other hand, utilizing event-based optical flow makes our method more robust to high-speed motions. Additionally, the performance of Ours\,(OF) method is much better compare to Ours\,(GE)~\cite{10342300}, which uses 1D brightness change as motion measurements and without prior information. This result demonstrates that 2D event-based optical flow can increase the motion observability with respect to 1D brightness change.

\subsection{Comparison with optical flow-aided object pose trackers}\label{comparison with optical flow-aided object pose trackers}

We also compare our method with ROFT~\cite{9568706}, which uses frame-based optical flow for object pose tracking. 
The performance of Ours\,(OF) is more consistent over a wider span of object speed than ROFT. 
ROFT performs well on regular motions. However, the accuracy of tracking deteriorates as the motion speed increases, as shown in Table~\ref{Main_table} and Fig.~\ref{cumulative_percentage_distribution}, because, as the object velocity increases, the accuracy of frame-based optical flow estimation is affected and decreases accordingly. 

\begin{figure*}[t]
    \centering
    \includegraphics[width=0.95
    \linewidth]{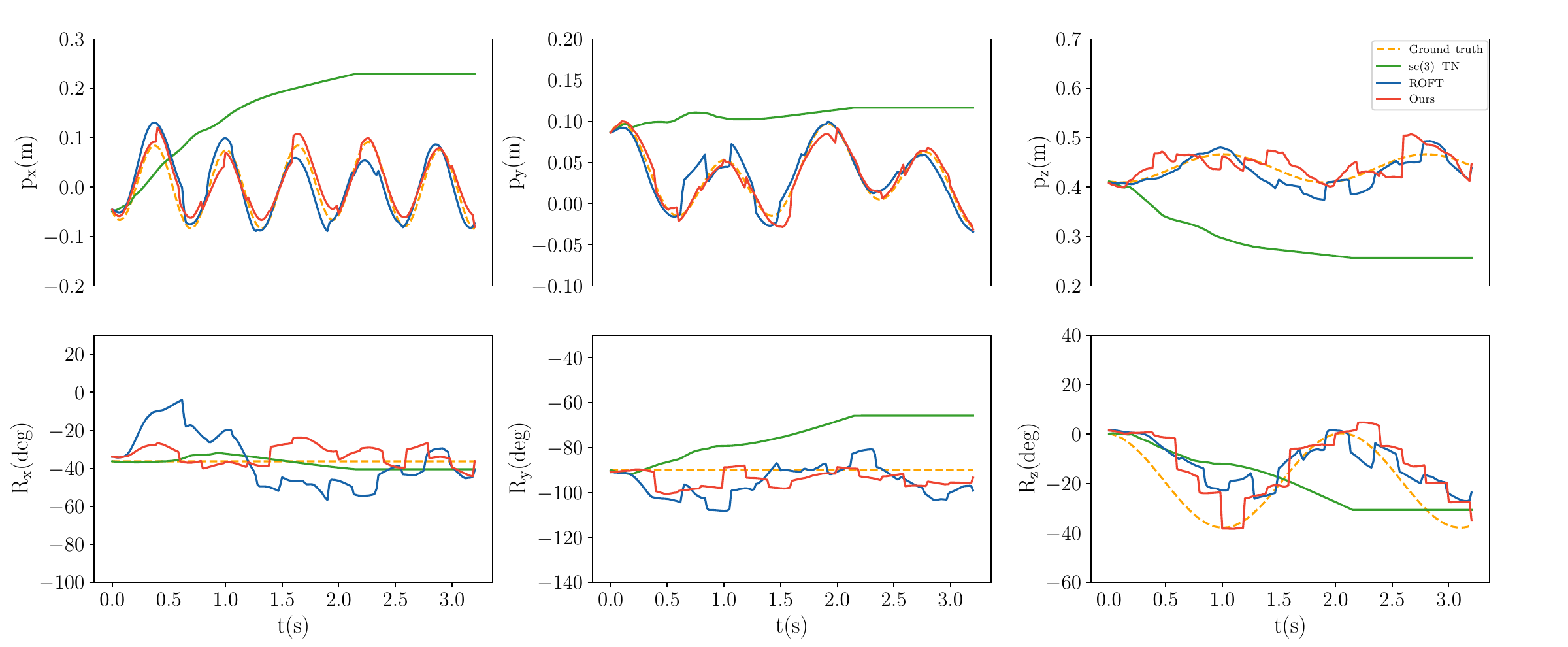}

    \caption{Comparison of the tracked object position and rotation generated by se(3)-TrackNet, ROFT, and Ours on the faster motion sample of 005\_Faster. Our method demonstrates higher accuracy.}   
    \label{tracking_curve}
\end{figure*}

In ROFT the same global object pose estimator DOPE~\cite{tremblay2018corl:dope} is used (at 5Hz), but it still fails on the faster motion sample 010\_potted\_meat\_can, due to the fact that ROFT adopts a pose outlier rejection strategy, and DOPE fails over time when motion blur becomes severe. In contrast, Ours\,(OF) can still track, showing that event-based optical flow is much more robust than frame-based optical flow when severe motion blur occurs. Additionally, the performance of Ours\,(OF) is much better compare to Ours\,(Pose), which only inputs DOPE estimated pose to the UKF. This shows the effectiveness of integrating event optical flow based object velocity tracker. A visualized tracking result of the sample 005\_Faster is reported in Fig.~\ref{tracking_curve}.

\begin{table}
    \vspace{0.7em}
    \renewcommand\tabcolsep{2.3pt}
	\centering
    \scriptsize
    \scalebox{1.0}{
    \begin{tabular}{c |c c| c c}
		\cline{1-5} 
        \toprule[1.1pt]
		\multicolumn{1}{c}{\textbf{Motion speed}}                            
		&\multicolumn{2}{c}{\textbf{Regular}}  
        &\multicolumn{2}{c}{\textbf{Faster}}\\
		\hline
        Methods &$e_{p}(cm)$ &$e_{r}(deg)$ &$e_{p}(cm)$ &$e_{r}(deg)$\\
		\hline
        se(3)-TrackNet  &7.69 &13.56  &38.88  &38.84\\
        ROFT      &\textbf{1.97} &9.15 &22.27 &36.60\\ 
        Ours\,(OF)  &2.53 &\textbf{8.82}  &\textbf{3.94}  &\textbf{11.56}\\ 
        \bottomrule[1.1pt]
	\end{tabular}}
        \caption{{Average tracking error of all testing sequences.}}
        \label{average_tracking_error}
\end{table}


    
    

\begin{figure}[htbp]
    \centering
        \hspace*{0.3cm}
        \includegraphics[height=7.6cm]{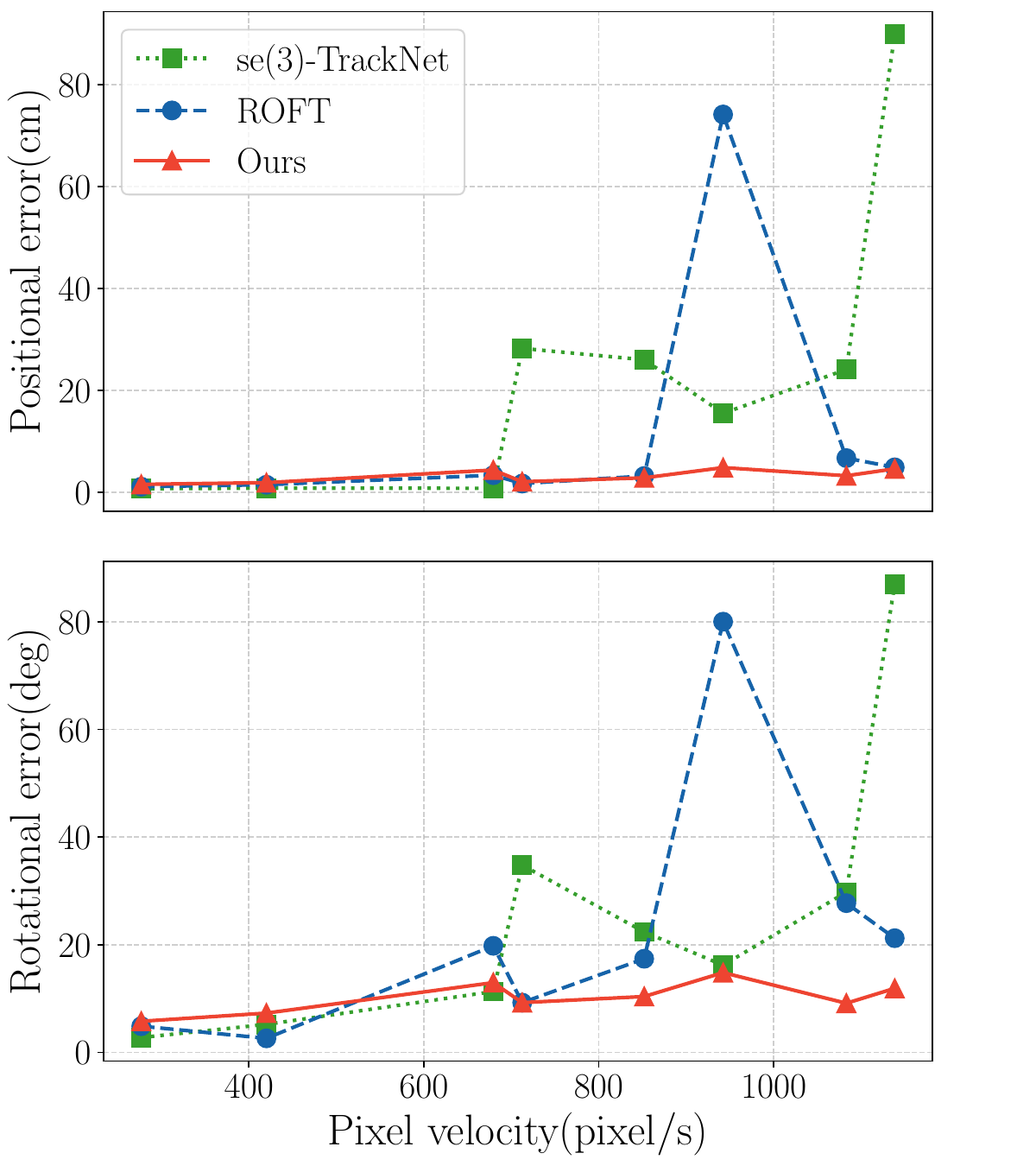}
    \caption{Tracking performance over pixel velocity.}    
\label{tracking_performance_over_pixel_velocity}
\end{figure}


\subsection{Testing on real-world data sequence}\label{realdatasets}
In the real world data testing, we compare our method Ours\,(OF) with se(3)-TrackNet and ROFT. Since no ground truth is available, only qualitative results are reported for this experiment. The tracked object pose is rendered with the object mesh and visualized on top of the collected images. By examining how the rendered object overlaps with the object in the frames, we can assess algorithms performance. 

\begin{figure*}[t]
    \centering
    \includegraphics[width=0.95
    \linewidth]{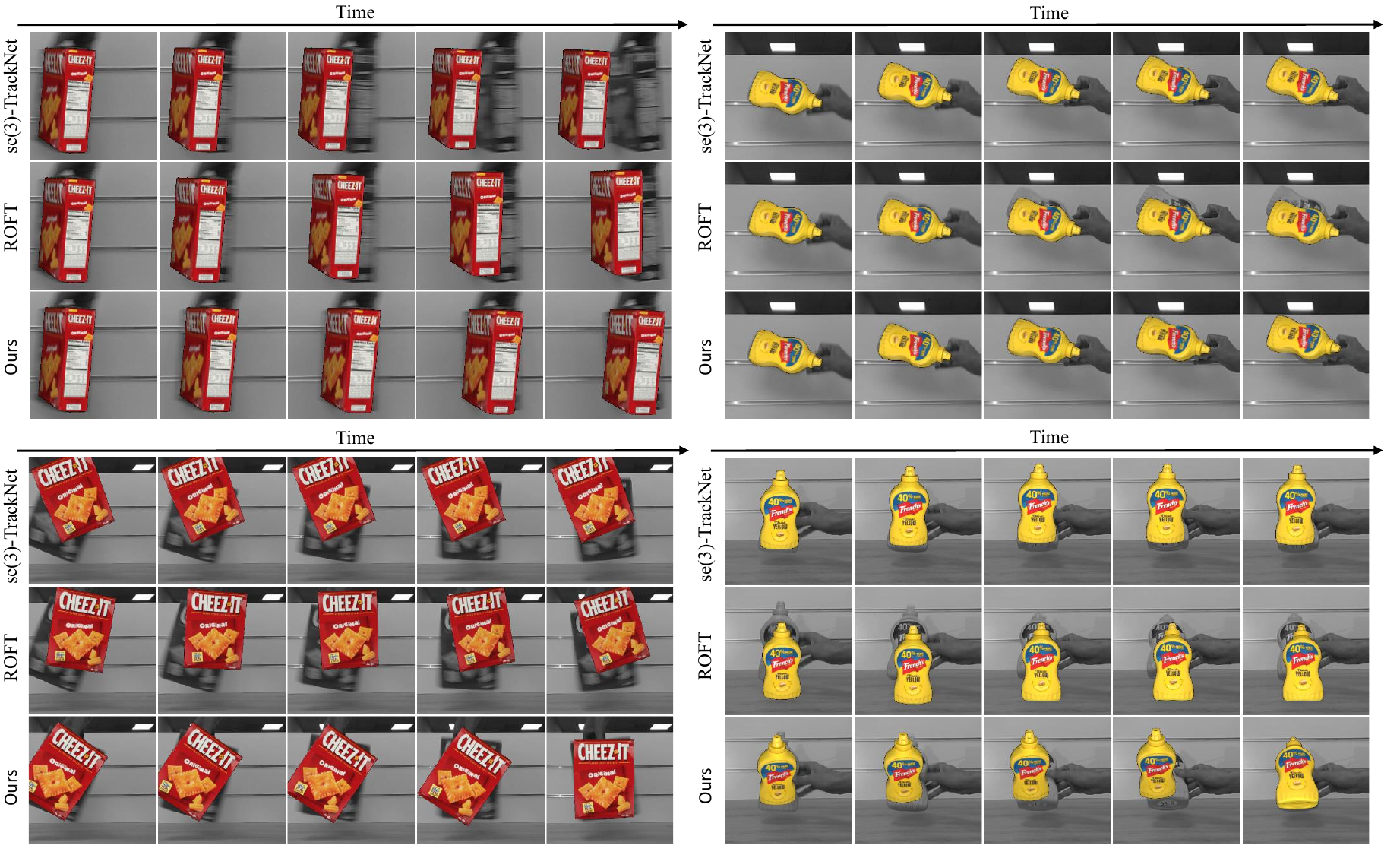}

    \caption{Qualitative results for 003\_cracker\_box and 006\_mustard\_bottle, include translation and rotation samples.}    
    \label{real_data_results}
\end{figure*}

Two translation samples 003\_cracker\_box translates along x-axis and 006\_mustard\_bottle translates along y-axis are tested for demonstrating the position tracking performance. Moreover, two rotation samples are tested, one is 003\_cracker\_box roll rotation and the other is 006\_mustard\_bottle pitch rotation. Noticeable motion blur exists in some of testing samples due to high-speed motion.  Fig.~\ref{real_data_results}, shows that se(3)-TrackNet works well for 006\_mustard\_bottle $y$ axis translation, but fails on 003\_cracker\_box $x$ axis translation, 006\_mustard\_bottle translation and 006\_mustard\_bottle pitch rotation.  These results are similar to the synthetic testing results, since high-speed motions means large jumps between two continuous frames, higher pixel velocity and more severe motion blur. ROFT is able to track real world samples, but the tracking error is very noticeable. Motion blur has impact on the estimation accuracy of the adopted frame-based optical flow, which is the same as reported in the simulated synthetic data sequences. Our method achieves relatively better tracking performance for translation samples, as the rendered object is almost overlapped with the one in the image. However, for rotation samples, tracking errors are very obvious. The reason is that rotations are not easy to be observed, especially with sparse measurements.

Differently from  se(3)-TrackNet, our tracker and ROFT have not encountered failures, thanks to the use of the global pose estimator DOPE. Overall, the results of real world testing are consistent with testing results on synthetic data.

\begin{table}
    \vspace{0.7em}
    \renewcommand\tabcolsep{2.3pt}
	\centering
    \scriptsize
    \scalebox{1.0}{
    \begin{tabular}{c |c c}
		\cline{1-3} 
        \toprule[1.1pt]
        \textbf{Methods} &$e_{p}(cm)$ &$e_{r}(deg)$ \\
		\hline
        w/o normal flow constraint  &4.93 &12.35\\
        w/o measurement weighting     &4.10 &10.45\\ 
        Ours  &\textbf{3.24} &\textbf{10.19}  \\ 
        \bottomrule[1.1pt]
	\end{tabular}}
    \caption{{Tracking performance of w/o normal flow constraint and measurement weighting.}}
    \label{ablation_study}
\end{table}

\subsection{Ablation study}
In this ablation study, we investigate the impact of normal flow constraint and measurement weighting on the tracking performance of our method on synthetic data. Results are reported in Table~\ref{ablation_study}.
\subsubsection{Normal flow constraint}
If normal flow constraint is removed, the RMSE of all testing sequences increases for both translation and rotation, demonstrating the existing impact of the aperture issue. By introducing the normal flow constraint, we can partially mitigate its effect.
\subsubsection{Measurement weighting}
The measurement weighting strategy applied to the measurements~(Equation.~\ref{eq:ekf_measurement_model}) helps eliminate some outliers, leading to a significant performance improvement.

\section{DISCUSSION}

All datasets had variations in velocity as the object would slow down to change direction. At these periods of zero velocity, motion blur is not present and the global pose estimation algorithm used by ROFT could recover from failure. se(3)-TrackNet could not recover from failure as it only iteratively estimates pose changes. Our algorithm achieved best results as the event optical flow was accurate even at high-speeds, which helped to offset errors of the global pose estimator. 

We ran the global pose estimator at a lower rate (5 Hz) than it could possibly operate (30 Hz with high-end GPU), demonstrating that the proposed approach can also save computational resources and reduce energy consumption.

To further increase robustness against motion blur, a global pose estimator could be developed that uses the event camera instead of the RGB-D camera. However, so far no algorithm exists and it is an open question of how to extract rich enough features from the event-stream for such a task.

Additionally, our tracking algorithm currently relies on the RGB-D camera to produce depth estimates as needed in the interactive matrix for 6-DoF velocity estimation. The development and integration of techniques for depth estimation without the RGB-D camera could allow for a fully event-based pipeline and also enable very low-latency asynchronous updates of 6-DoF velocity.

The current experiments were performed offline. We are confident that the optical flow and the Kalman filters are fully real-time capable and an extended experimentation could test the latency and stability in live conditions required by intelligent robots.

\section{CONCLUSIONS} 
In this work, we propose using event-based optical flow as motion measurement, combining with low frequency global estimated pose for object pose tracking for high-speed motion scenarios. A Kalman filter and an Unscented Kalman filter, are adopted for object 6-DoF velocity tracking and 6-DoF pose tracking, respectively. Comparison against a state-of-the-art CNN-based object pose tracker and a frame-based optical flow aided tracker demonstrates the effectiveness of integrating event-based optical flow for object pose tracking especially for high-speed motion scenarios. This is confirmed both by simulated and real work data, demonstrating the practical application of the proposed system.

\addtolength{\textheight}{-9.8cm}   









\bibliography{root}

\end{document}